\documentclass[10pt,twoside,twocolumn]{article}
\usepackage{blindtext} 
\usepackage[sc]{mathpazo} 
\usepackage[T1]{fontenc} 
\usepackage{float}
\usepackage{microtype} 
\usepackage[english]{babel}
\usepackage[hmarginratio=1:1,top=20mm,columnsep=20pt]{geometry} 
\usepackage[hang, small,labelfont=bf,up,textfont=it,up]{caption}
\usepackage{booktabs} 
\usepackage{lettrine} 
\usepackage{graphicx}
\usepackage[hidelinks]{hyperref}
\usepackage{caption}
\usepackage{subcaption}
\usepackage{enumitem}
\setlist[itemize]{noitemsep}
\usepackage{abstract}

\usepackage{titlesec} 
\renewcommand\thesection{\Roman{section}} 
\renewcommand\thesubsection{\roman{subsection}} 
\titleformat{\section}[block]{\large\scshape\centering}{\thesection.}{1em}{} 
\titleformat{\subsection}[block]{\large}{\thesubsection.}{1em}{} 
\usepackage{fancyhdr} 
\usepackage{titling}
\usepackage{hyperref}
\usepackage{cite}
\pretitle{\begin{center}\Huge\bfseries}
\posttitle{\end{center}} 
\title{Improving on Q \& A Recurrent Neural Networks Using Noun-Tagging}
\author{%
	 \textsc{Omar El-lakany} \hspace{1cm} \textsc{Erik Partridge}\hspace{1cm} \textsc{Jack Sklar} \\
     \normalsize \href{mailto:omar.ellakany@mail.mcgill.ca}{omar.ellakany@mail.mcgill.ca}
     \normalsize \href{mailto:erik.partridge@mail.mcgill.ca}{erik.partridge@mail.mcgill.ca}
	\normalsize \href{mailto:jack.sklar@mail.mcgill.ca}{jack.sklar@mail.mcgill.ca}\\
    \normalsize McGill University
}
\date{July 12, 2018} 
\renewcommand{\maketitlehookd}{%
	\begin{abstract}
		\noindent Often, more time is spent on finding a model that works well, rather than tuning the model and working directly with the dataset. Our research began as an attempt to improve upon a simple Recurrent Neural Network for answering "simple" first-order questions (QA-RNN), developed by Ferhan Ture and Oliver Jojic, from Comcast Labs \cite{Ture}, using the SimpleQuestions dataset. Their baseline model, a bidirectional, 2-layer LSTM RNN and a GRU RNN, have accuracies of $0.94$ and $0.90$, for entity detection and relation prediction, respectively. We fine tuned these models by doing substantial hyper-parameter tuning, getting resulting accuracies of $0.70$ and $0.80$, for entity detection and relation prediction, respectively. An accuracy of $0.984$ was obtained on entity detection using a 1-layer LSTM, where preprocessing was done by removing all words not part of a noun chunk from the question. 100\% of the dataset was available for relation prediction, but only $20\%$ of the dataset, was available for entity detection, which we believe to be much of the reason for our initial difficulties in replicating their result, despite the fact we were able to improve on their entity detection results.
	\end{abstract}
}

\begin{document}
\maketitle

\section{Introduction}

\lettrine[nindent=0em,lines=2]{F}irst-order factoid question answering (QA) is a text classification problem of being able to answer a question which is based on a single trivial answer that is assumed to belong in a knowledge base (KB). Approaches usually employ manually defined string matching rules, bag of words representations, or use of linguistic heuristics, which restrict the domain in which their classifiers are useful. The use of an Recurrent Neural Network (RNN), on the other hand, allows for high accuracy simple-question answering, which only require a Q \& A set with labeled question subjects (entities) and labeled subject-object relationships, and requiring little to no prior linguistic knowledge. 

In general, a more complex method might need to be used if the assumption can not be made, that the question types are first-order factoid questions, and that the fact needed to answer such questions belong in our KB. Since we can make that assumption, a more simple approach has been shown to work better \cite{Ture}, although we show that it is possible to further simplify the approach which lead to even further improvements to the accuracy of a QA-RNN on entity detection. We propose that with further hyper-parameter tuning of a less complex model compared to the simple baseline model given in Ture et al. \cite{Ture}, that the accuracy on the SimpleQuestions dataset can be further improved in entity detection, given the assumption made that the vast majority of the questions are grammatical.

\section{Simple Question Answering Approach}

The assumption is made that the answer to any first-order question is a single property of a single entity in the KB. Two important parts of a question define how a search for the fact is performed in the KB. Entities of a question, the question's subject, are words that are indicative of the answer of that question, and therefore are the most significant words in the question. Relations are defined as the relation between the subject of the question and the answer. In an attempt to generalize to broader QA examples and to reduce complexity, they enforced stricter assumptions on the problem structure.

The Knowledge base contains a very-large set of facts, each representing a binary relationship of the form: [subject, relation, object]. Due to the first-order question assumption, only a single fact from the KB is needed to answer any single question. The Question, for example, "How old is Tom Hanks?", is reduced from its original form, to a structured query of the form: \{entity: Tom Hanks, relation: bornOn\}. This type of preprocessing is performed in two tasks: entity detection, tagging each word in the input question using binary representation as either an entity word or not, and relation prediction, classifying the question into one of K different relation-classes. Both steps were approached by our adversaries using an RNN.

A standard RNN architecture is used: each question word is transformed from a one-hot-encoding to a d-dimensional vector, using Google News Word2Vec. Then a Recurrent layer combined the vectorized input with hidden layer representation from the previous word and applies a set of non-linear activation functions to compute hidden layer representation for the current input word. Soft-max is applied to the final hidden representation of the final recurrent layer to obtain a normalized probability distribution, and is returned as output. For entity detection, words are either classified as an entity ($= 1$) or as context ($= 0$). For relation prediction, the output is one of 1837 relation-classes.

Entity phrases are constructed by concatenating consecutive entity words returned from the entity-RNN, and then the structured-query is constructed using the entity phrase and the predicted relation class. In order to match the predicted-entity with the true-entity in the KB, our adversaries build an inverted index, $I_{\rm entity}$. $I_{\rm entity}$ maps  n-grams of size, 1,2, and 3, to the entities alias text, with associated TF-IDF (Term Frequency-Inverse Document Frequency). Once $I_{\rm entity}$ is constructed, the returned elements from the KB that match the entity phrase constitute a set of candidate entities, denoted as $C$. A graph reachability index, $I_{\rm reach}$, is built that maps each entity node, $e_i \in C$, to all nodes in the KB that have matching entities, $e_i$. $I_{\rm reach}$ is used to find any nodes that have the relation class that correspond to the relation class predicted using the Relation-RNN. These nodes constitute a candidate-answer set, denoted $A$. Once this is done for each candidate entity node in $C$, the node with the highest corresponding score in $A$ is returned as the most likely answer.

\section{Baseline Experimental Setup}

The SimpleQuestions dataset \cite{Bordes}, was used by Ture et al., which contained 108,442 questions annotated by hand in English, from a subset of knowledge from the FreeBase dataset\footnote{The FreeBase dataset is a large knowledge base of 17.8 million facts, on 4 million unique entities, consisting 7523 relation types.}. The questions were randomly shuffled and 70\% of them were used as a training set, 10\% as a validation set, and the remaining 20\% as test set. A pre-trained embedding layer, GoogleNews Word2Vec, was used to embed the words into 300-dimensional vectors. Parameters of their RNNS were learned via stochastic gradient descent, with categorical cross-entropy as the error function. The maximum size of of the input length, $N$, was set to 36, and a special character was used if the input was shorter than $N$. 

Ture et al. tested 48 RNN model configurations based on, 4 different types of RNN layer (GRU or LSTM, bidirectional or not), the layer depths ($1,2,3$), and and the drop-out ratio ($0$ to $0.5$). Their optimal model configuration, and our baseline model, determined by the optimal accuracy on the validation set, was a LSTM (BiLSTM2) for entity detection and a GRU (BiGRU2) for relation prediction, both being bidirectional, 2-layer RNNs with drop-out ratios of 10\%.

Ture et al. followed the metric of Bordes et al. in their method for defining if a question is correctly answered, that being if and only if the entity selected, $e$, and the relation predicted, $r$, match the true entity and relation, respectively. This is based off of the assumption, that if both the entity and relation are predicted correctly, given the KB, the correct answer must be found, when searched for using $e$ and $r$.

\begin{table*}[!htp]
\begin{center}
\caption{Results of all Classifiers on both entity detection and relation prediction of the SimpleQuestions Dataset, as well as results of the QA-RNN developed by Ture et al. The results of the classifiers NT-QA, Multinomial NB, Random Forests, Bernoulli NB, and KNN, all employed the noun-tagging preprocessing step for entity detection.}
\begin{tabular}{|c||c|c|} 
 \hline
Classifier & ED Accuracy & RP Accuracy \\ 
 \hline
NT-QA (Ours) & 0.984 & N/A \\
QA-RNN (Ture et al.) & 0.94 & 0.90 \\
QA-RNN (tuned) & 0.70 & 0.80 \\
QA-RNN (simplified) & 0.72 & 0.79 \\
Multinomial NB & 0.77 & 0.59 \\
Random Forests & 0.85 & 0.56 \\
Bernoulli NB & 0.78  & 0.59 \\
SGD & N/A  & 0.61   \\
KNN & 0.83 &  0.56 \\
Voting (Bernoulli, KNN, SGD) & N/A  & 0.60  \\
 \hline
\end{tabular}\label{tab:1}
\end{center}
\end{table*}

\section{Results}

We were unable to utilize the full SimpleQuestions dataset for the entity detection, as a result of the dataset no longer being as available as it was previously, so we made due with roughly $~20\%$ of the dataset, but 100\% of the SimpleQuestions dataset was used for relation prediction. As a subset of the FreeBase dataset, the SimpleQuestions dataset has tagged entities saved using FreeBase ID \#'s. The problem is that the FreeBase API is no longer in existence, so the Entity ID \#'s can not be decoded. $20\%$ of the SimpleQuestions dataset, the test set from Ture et al., was found online with both Relations and Entities decoded. The Relation ID \#'s were already decoded, so it was possible to use the entire SimpleQuestions dataset to train our relation prediction QA-RNN. The baseline models of Ture et al., BiLSTM2 for ED and BiGRU2 for RP, were reproduced with extensive hyper-parameter tuning, with the only difference being that the slim version of the GoogleNews Word2Vec embedding layer was used.

Next, the baseline model used by Ture et al. was simplified in its complexity, in an attempt to improve on the baseline performance. A set of simple classifiers were trained and tested on the SimpleQuestions dataset, with an added preprocessing step used for the entity detection task. Being as none of these simple classifiers have any type long short-term memory mechanisms as with LSTMs, it was hypothesized that the accuracy of these classifiers would not beat the accuracy of the QA-RNN defined in Ture et al.

\subsection{Reproducing Existing Results}
To replicate the model of Ture et. al's model, both the BiGRU2 and BiLSTM2 were created using Keras. Then, hyper-parameter tuning was performed for activation regularization, learning rate, and neuron count (dropout and activation function were both defined by the paper).

A basin-hopping algorithm was used on each network, spanning approximately 150 models tested for each network. A neuron ratio from first to second hidden layer of the BiGRU2 was optimized at 3.5, with a minimum of 400 neurons, and a ratio of 3.1 for the BiLSTM2 with a minimum of 400 neurons. L1 regularization on the activity of each layer of both networks at a rate of 0.01 was found to be optimal in both cases.

The hyper-parameter tuning was done using the Adam learning algorithm (although the Adam algorithm used did unfortunately fall prey to the implementation error noted by Loschilov et. al\cite{Loschilov}), using categorical cross-entropy. Due to performance challenges, stochastic gradient descent, the algorithm defined in the paper, was only used once the network was tuned, and then was tuned for learning rate. However, in neither case was the network able to replicate the results listed by Ture et. al..

\subsection{Simplifying and Ameliorating the Approach}

In our efforts to find a better method, we implemented a wide range of machine learning models. In both entity detection and relation prediction, simple non-neural models were tested. While no non-neural model performed well on entity detection, several performed reasonably well on relation prediction. However, these did not perform well when combined or kerneled, and are unlikely to perform much better than the listed accuracies, which remain substantively below those of the neural models.

While their model performs decently well in relation prediction, having 2 bidirectional GRU layers leads to a complex model that requires a lot of training time. In order to solve this issue, we designed a simpler implementation of their model by removing one of the GRU layers, and replacing it with a 1D convolutional layer. This layer can identify patterns in the text through convolutional operations and we also placed this layer below the embedding layer to provide better embeddings. This is a simpler model as bi-directional GRU's are very complex layers and incur substantial computational cost. By removing one of the layers and replacing it with a simple convolutional layer, we reduce the training time by about $40\%$. This result draws upon some of the power of convolutional networks in computer vision and exemplifies their relevance to natural language processing, although it may be a greater comment upon the structure of the dataset than on convolutional networks' effectiveness in natural language as a whole.

To optimize our model, we performed hyper-parameter tuning using a similar approach to the optimization of our implementation of their model with a minimum of 400 neurons as well for the GRU layer in relation detection. For the convolutional layer, we required 50 different filters with a small window size of 2 being optimal  and we also used a dropout with a ratio of 0.2 after the convolutional layer and 0.1 after the GRU layer. Finally, we determined that even a flawed Adam \cite{Loschilov} was the best optimization function with a learning rate of $0.0007$ and that the best loss function for relation detection is categorical cross-entropy. This model obtained an accuracy of about $79\%$ which was poor in comparison to the results of Ture et al., but reasonable for the performance boost that it gave.

Our improvements on the results of entity detection of Ture et al. were greatest when using a form of natural language preprocessing using SpaCy. First, we tagged the question components as to if they were part of a noun-cluster as defined by SpaCy. Then, because in a grammatically correct question, the subject (entity) should be a noun, we removed everything that was not part of a noun-cluster. We then trained a single-layer bidirectional LSTM on that data, and achieved an accuracy of $98.4\%$ on grammatically correct questions for entity detection, which was a significant improvement, while reducing the complexity of the model. A preprocessing step for relation prediction (removing all words that are not nouns, verbs, or adverbs from the input to relation prediction) did not work as well, and was not used. 

\section{Discussion}

The decision by Ture et al., not to focus on attention mechanisms or linguistic prior knowledge, but rather to focus on what type of model best takes advantage of the problem structure, is what makes the performance of their QA-RNN classifier far better than that of other attempts made around the same time. Although, the decision not tag the input question for part-of-speech before inputting to the entity detection QA-RNN, was a major shortcoming of their approach. Given that the subject of a question, the entity, should almost always be a noun, it is a very easy preprocessing step to filter out all non-nouns from the input of the entity QA-RNN , which increased the performance on the test set from $70\%$ to $98\%$, a major improvement.

Relation prediction is very critical, shown by a test of Ture et al.: using a naive majority classifier, which classifies all relations to be the most common class, "BornOn", performed with an accuracy of $4.1\%$. On the other hand, the naive entity detector tested by Ture et al., where the entire question is tagged as an entity, performed with an accuracy of $58.9\%$. This shows how significant the relation prediction is to finding the correct factoid in the Knowledge Base.

One major shortcoming of the approach of Ture et al. is the fact that they used Stochastic Gradient Descent (SGD), rather than another approach such as the Adam or Adamax optimizer, which showed substantially better performance than SGD. We attempted to improve upon their approach, by using different optimization functions, Adam being the best, which in theory does not add any complexity to our model. Another shortcoming of their approach is their failure to use any form of regularization on the activation of each layer. Adding L1 regularization to each layer showed a small, but noticeable, improvement to the results.

It should also be noted that the full dataset would be much more important in terms of training our entity detection QA-RNN (BiLSTM2), rather than the relation prediction QA-RNN (BiGRU2). For one, this is because the ED QA-RNN is an LSTM, which is a more complex model than a GRU, so more data is needed to train. Also, with a maximum question length of $36$, the task of entity detection has a upper-bound of $2^{36}$ ways of tagging a question, compared to relation prediction having a total $1837$ classes to predict from. The preprocessing step of removing all words except for nouns, using the SpaCy\cite{Spacy} Natural Language Processor, for entity detection, reduced the vocabulary size, and therefore the complexity of the embedding layer, by a factor of $10$, which drastically increased our accuracy on their model for entity detection. 

Our results using the hyper-parameter tuned QA-RNN model from Ture et al., an accuracy of $0.80$, are significantly worse than the results reported in Ture et al. ($0.90$). With more resources for hyper-parameter tuning for our relation prediction QA-RNN, we believe that we would come closer to reproducing their results. Performing some form of optimization based on preprocessing the question word might help reduce the search space, as would additional filtering based on verbs. One challenge that was encountered in this setting, is that some classes for relation prediction only have one entity---this makes it very difficult for a network to learn without using a one-shot training algorithm. As such, integrating a one-shot training algorithm, or gathering a larger dataset, would likely to improve performance.

\appendix
\section{Software}
The Keras\cite{keras} Python library was used to implement all RNNs mentioned in this paper. SKLearn\cite{sklearn} was used to implement naive classifiers and SciPy\cite{scipy} was used for hyper-parameter tuning. GenSim was used in order to obtain the pre-trained Google News Word2Vec \cite{Gensim} and SpaCy\cite{Spacy} was used to tag nouns for the entity detection task. 

\section{Acknowledgements}
We would like to thank Ferhan Ture for his help in troubleshooting our attempts to gather the entire SimpleQuestions dataset. 
\bibliographystyle{plain}
\bibliography{citations}

\end{document}